\documentclass{article}

\usepackage[final]{neurips_2023}

\usepackage[utf8]{inputenc}
\usepackage[T1]{fontenc}
\usepackage{hyperref}
\usepackage{url}
\usepackage{booktabs}
\usepackage{amsfonts}
\usepackage{amsmath}
\usepackage{nicefrac}
\usepackage{microtype}
\usepackage{xcolor}
\usepackage{graphicx}
\usepackage{algorithm}
\usepackage{algorithmic}
\usepackage{enumitem}
\usepackage{tikz}
\usetikzlibrary{arrows.meta, positioning, shapes.geometric, calc, decorations.pathreplacing}
\usepackage{listings}
\usepackage{multirow}
\usepackage{pifont}
\newcommand{\cmark}{\ding{51}}
\newcommand{\xmark}{\ding{55}}

\definecolor{keepgreen}{RGB}{34,139,34}
\definecolor{revertred}{RGB}{178,34,34}
\definecolor{stageblue}{RGB}{30,80,150}

\title{AutoKernel: Autonomous GPU Kernel Optimization\\via Iterative Agent-Driven Search}

\author{
  Jaber Jaber\thanks{Correspondence: \texttt{jaber@rightnowai.co}} \\
  RightNow AI\\
  \texttt{jaber@rightnowai.co} \\
  \And
  Osama Jaber \\
  RightNow AI\\
  \texttt{osama@rightnowai.co} \\
}

\begin{document}

\maketitle

\begin{center}
\includegraphics[height=1.1cm]{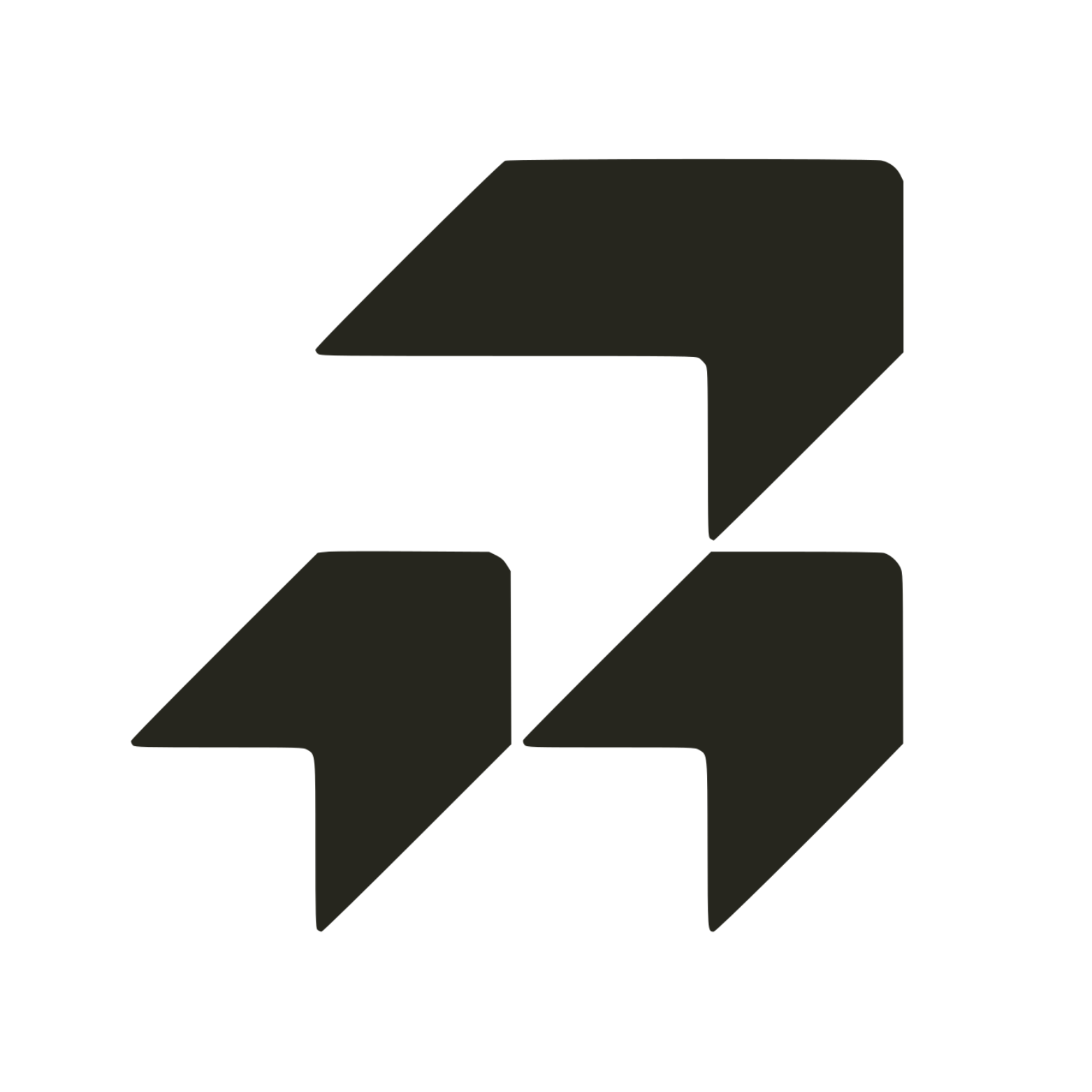}
\end{center}

\vspace{-0.3em}

\begin{abstract}
Writing high-performance GPU kernels is among the most labor-intensive tasks in machine learning systems engineering. A single matrix multiplication kernel targeting tensor core hardware may require weeks of expert tuning across tiling strategies, memory layouts, and precision configurations. We present \textsc{AutoKernel}, an open-source framework that applies an autonomous agent loop to GPU kernel optimization for arbitrary PyTorch models. Given a model, \textsc{AutoKernel} profiles it to identify computational bottlenecks, ranks them by Amdahl's law impact, and iteratively refines Triton or CUDA~C++ kernel implementations through hundreds of experiments without human intervention. A five-stage correctness harness, covering smoke tests, shape sweeps across 10+ configurations, numerical stability under adversarial inputs, determinism verification, and edge-case coverage, ensures that every candidate kernel is validated before any speedup is recorded. We describe the system architecture comprising over 9{,}000 lines of Python and a 909-line agent instruction document, 18 starter kernel implementations across two backends, a six-tier optimization playbook, and integration with the KernelBench benchmark suite of 250 standardized problems. \textsc{AutoKernel} covers nine kernel types spanning the dominant operations in modern transformer architectures. On an NVIDIA H100, our Triton kernels outperform both PyTorch eager and \texttt{torch.compile} (\texttt{max-autotune}) on the majority of tested configurations: 5.29$\times$ over eager on RMSNorm, 2.82$\times$ on softmax, and 2.21$\times$ on cross-entropy, while beating \texttt{torch.compile} by 2.83$\times$, 3.44$\times$, and 2.94$\times$ on the same kernels respectively. In community deployment, an AutoKernel-optimized kernel achieved first place on the \texttt{vectorsum\_v2} B200 leaderboard, and a single-prompt Triton FP4 matmul kernel was reported to outperform CUTLASS by 1.63 to 2.15$\times$. The full system is available at \url{https://github.com/RightNow-AI/autokernel}.
\end{abstract}

\section{Introduction}

The runtime of large transformer models on GPU hardware is dominated by a small set of computational kernels. Matrix multiplications in attention and feed-forward layers typically consume 60 to 80\% of total GPU time, with normalization, softmax, and positional embedding operations accounting for most of the remainder. Vendor libraries like cuBLAS and cuDNN cover common cases well, but the pace of architectural innovation in deep learning regularly outstrips library coverage: grouped-query attention, SwiGLU activations, rotary position embeddings, and RMS normalization all shipped in production models before receiving dedicated library support.

Closing the gap between default library performance and hardware capability demands deep expertise in GPU microarchitecture. The engineer must reason about arithmetic intensity and the roofline model~\citep{williams2009roofline}, memory coalescing and bank conflicts in shared memory, register pressure and occupancy tradeoffs, tile sizes and their interaction with the L2 cache, warp-level synchronization, and tensor core instruction selection. A single high-performance matmul kernel may involve 200+ lines of CUDA or Triton code with dozens of interdependent parameters. This expertise is scarce, and the tuning process scales poorly.

\paragraph{Can LLMs write GPU kernels?} KernelBench~\citep{ouyang2025kernelbench} posed this question systematically, evaluating frontier LLMs on 250 GPU kernel problems across four difficulty levels. The answer was sobering: even the best models matched PyTorch baseline performance in fewer than 20\% of cases using one-shot generation. Subsequent work has explored multi-agent systems~\citep{astra2025, kernelskill2026}, evolutionary search~\citep{kernelfoundry2026, gpukernelscientist2025}, hardware-aware feedback loops~\citep{cudaforge2025}, and reinforcement learning~\citep{cudal1, cudaagent2026} to push these numbers higher.

\paragraph{Our approach.} Rather than engineering a complex multi-agent architecture, we observe that the workflow of an expert kernel engineer is itself a simple loop: write a candidate, benchmark it, keep improvements, discard regressions, repeat. \textsc{AutoKernel} mechanizes this loop. An LLM agent modifies a single kernel file, a fixed benchmark harness verifies correctness and measures throughput, and the result determines whether the change persists. Each iteration takes roughly 90 seconds. An overnight run produces 300 to 400 experiments across multiple kernels.

This design draws directly from Karpathy's \texttt{autoresearch} project~\citep{karpathy2026autoresearch}, which demonstrated that an AI agent running a keep/revert loop on LLM training code could discover 20 optimizations across 700 experiments in two days on a single GPU. AutoKernel transplants this loop to kernel code: the search space becomes the space of possible kernel implementations, and the evaluation function is a correctness-gated benchmark rather than validation loss.

\paragraph{Key insight: optimize what matters.} Unlike prior work that treats kernel problems in isolation, \textsc{AutoKernel} operates on complete models. It profiles an arbitrary PyTorch model, identifies which kernels consume the most GPU time, and applies Amdahl's law to allocate optimization effort proportionally. A 1.5$\times$ speedup on a kernel that accounts for 60\% of total runtime yields 1.25$\times$ end-to-end. The same speedup on a 5\% kernel yields only 1.03$\times$. The system focuses effort where it compounds.

\paragraph{Contributions.}
\begin{enumerate}[leftmargin=*, topsep=2pt, itemsep=1pt]
\item A complete, open-source pipeline (9{,}200+ lines of Python, plus agent instructions) for autonomous GPU kernel optimization, from model profiling through end-to-end verification.
\item A five-stage correctness harness testing 10+ input shapes, three data types, adversarial inputs, determinism, and non-power-of-two edge cases.
\item A dual-backend architecture with 9 Triton and 9 CUDA~C++ starter kernels, covering the dominant operations of transformer models.
\item An Amdahl's-law orchestrator with explicit move-on criteria (plateau detection, peak utilization, time budget, speedup threshold).
\item A six-tier optimization playbook encoding expert kernel engineering knowledge into agent-readable instructions.
\item Integration with the KernelBench benchmark suite and HuggingFace Kernels distribution platform.
\end{enumerate}

\section{Related Work}

\subsection{GPU Kernel Languages and Compilers}

Triton~\citep{tillet2019triton} introduced tile-based GPU programming where the developer reasons about block-level tensor operations rather than individual threads. The compiler handles memory coalescing, shared memory allocation, and instruction scheduling, enabling FP16 matmul kernels that match cuBLAS in under 25 lines of code. CUDA~C++~\citep{nvidia2024cuda} remains necessary for explicit access to warp-level primitives, tensor core instructions (WMMA/MMA), and shared memory bank-conflict-free layouts. PyTorch~2's TorchInductor~\citep{ansel2024pytorch2} compiles captured computation graphs into Triton kernels automatically, achieving 2.27$\times$ inference and 1.41$\times$ training speedup across 180+ models. AutoKernel supports both Triton and CUDA~C++: Triton for rapid iteration (1 to 5 second compilation), CUDA~C++ for maximum control.

\subsection{Optimized Kernel Libraries}

FlashAttention~\citep{dao2022flashattention} demonstrated that tiled attention with online softmax can reduce HBM reads and writes from quadratic to linear in sequence length, achieving 2 to 4$\times$ wall-clock speedups. FlashAttention-2~\citep{dao2023flashattention2} reached 50 to 73\% of theoretical peak on A100. FlashAttention-3~\citep{shah2024flashattention3} exploited Hopper asynchrony (TMA, WGMMA) and FP8 quantization to reach 840~TFLOPS on H100 (85\% utilization). CUTLASS~\citep{thakkar2023cutlass} provides composable tensor core GEMM building blocks. These libraries represent the performance ceiling that AutoKernel's agent works toward.

\subsection{LLM-based Kernel Generation}

Table~\ref{tab:related} compares recent systems along several axes.

\begin{table}[t]
\centering
\caption{Comparison of LLM-based GPU kernel optimization systems. \textsc{AutoKernel} is the only system that combines model-level profiling with dual Triton/CUDA~C++ backend support.}
\label{tab:related}
\footnotesize
\setlength{\tabcolsep}{2.5pt}
\begin{tabular}{@{}lccccc@{}}
\toprule
\textbf{System} & \textbf{Method} & \textbf{Backend} & \textbf{Model} & \textbf{Iterative} & \textbf{Open} \\
 & & & \textbf{prof.} & & \textbf{src} \\
\midrule
KernelBench~\citep{ouyang2025kernelbench} & Benchmark & Multi-DSL & \xmark & \xmark & \cmark \\
GEAK~\citep{geak2025} & Multi-agent & Triton & \xmark & \cmark & \cmark \\
CudaForge~\citep{cudaforge2025} & Coder+Judge & CUDA & \xmark & \cmark & \cmark \\
KernelFoundry~\citep{kernelfoundry2026} & Evolutionary & CUDA/SYCL & \xmark & \cmark & \cmark \\
Kernel Scientist~\citep{gpukernelscientist2025} & Evolutionary & CUDA & \xmark & \cmark & \cmark \\
CUDA-L1~\citep{cudal1} & RL & CUDA & \xmark & \cmark & \cmark \\
CUDA Agent~\citep{cudaagent2026} & Agentic RL & CUDA & \xmark & \cmark & \cmark \\
Astra~\citep{astra2025} & Multi-agent & CUDA & \xmark & \cmark & \cmark \\
KernelSkill~\citep{kernelskill2026} & Multi-agent & CUDA & \xmark & \cmark & \cmark \\
\midrule
\textbf{AutoKernel} & Iterative loop & Triton+CUDA & \cmark & \cmark & \cmark \\
\bottomrule
\end{tabular}
\end{table}

KernelBench~\citep{ouyang2025kernelbench} provides the standard evaluation infrastructure. GEAK~\citep{geak2025} introduces generation, evaluation, reflection, and optimization agents for Triton kernels on AMD hardware. CudaForge~\citep{cudaforge2025} pairs a coder with a hardware-aware judge that provides profiling feedback. KernelFoundry~\citep{kernelfoundry2026} applies MAP-Elites quality-diversity search with meta-prompt evolution, reporting 2.3$\times$ average speedup on KernelBench for SYCL. GPU Kernel Scientist~\citep{gpukernelscientist2025} formulates optimization as multi-stage evolution on AMD MI300. CUDA-L1~\citep{cudal1} combines supervised fine-tuning with contrastive RL, achieving 3.12$\times$ average speedup and acceptance at ICLR 2026. CUDA Agent~\citep{cudaagent2026} applies large-scale agentic RL, achieving 100\% faster-than-\texttt{torch.compile} rate on KernelBench Levels 1 and 2, and 92\% on Level~3. Astra~\citep{astra2025} coordinates specialized profiling and planning agents, reporting 1.32$\times$ average speedup. KernelSkill~\citep{kernelskill2026} introduces dual-level memory with reusable expert optimization skills, reaching 5.44$\times$ on Level~1.

AutoKernel differs from these systems in three respects. First, it starts from a complete PyTorch model, profiles it, and focuses optimization on kernels ranked by their contribution to total runtime. Second, it supports both Triton and CUDA~C++ backends within a single framework, letting the agent choose the right abstraction level per kernel. Third, it uses a simple loop (edit, benchmark, keep/revert) with a five-stage correctness harness rather than multi-agent coordination or learned policies, trading architectural complexity for transparency and reliability.

\subsection{Autonomous Research Agents}

Karpathy's \texttt{autoresearch}~\citep{karpathy2026autoresearch} demonstrated autonomous LLM training optimization: an agent modifies a 630-line training script, runs a fixed 5-minute evaluation, and commits improvements. It ran 700 experiments in two days on a single GPU, discovering 20 training optimizations without human input. AutoKernel applies this paradigm to kernel optimization, with a different search space and evaluation function.

\section{System Design}

\begin{figure}[t]
\centering
\begin{tikzpicture}[
  box/.style={draw, rounded corners=3pt, thick, minimum height=0.65cm, align=center, font=\footnotesize\sffamily, text width=2.0cm},
  arr/.style={-{Stealth[length=4pt]}, thick},
  looparr/.style={-{Stealth[length=4pt]}, thick, orange!70!black},
]

\node[box, fill=blue!10] (model) at (0, 0) {PyTorch\\Model};
\node[box, fill=blue!10] (profile) at (0, -1.2) {Profiler};
\node[box, fill=blue!10] (extract) at (0, -2.4) {Extractor};

\node[box, fill=orange!12] (agent) at (3.2, -0.6) {Agent edits\\kernel.py};
\node[box, fill=orange!12] (bench) at (3.2, -1.8) {5-Stage\\Benchmark};
\node[box, fill=orange!12] (orch) at (3.2, -3.0) {Orchestrator};

\node[box, fill=green!10] (verify) at (6.2, -1.8) {End-to-End\\Verifier};

\draw[arr] (model) -- (profile);
\draw[arr] (profile) -- (extract);

\draw[arr] (extract.east) -- ++(0.4,0) |- (agent.west);

\draw[arr] (agent) -- (bench);
\draw[arr] (bench) -- node[right, font=\scriptsize] {result} (orch);

\draw[looparr] (orch.west) -- ++(-0.7,0) |- node[left, font=\scriptsize, pos=0.25] {continue} (agent.west);

\draw[arr] (orch.east) -- ++(0.4,0) |- node[above, font=\scriptsize, pos=0.7] {done} (verify.west);

\node[font=\scriptsize\bfseries\sffamily, blue!60!black, anchor=south] at (0, 0.5) {Phase A};
\node[font=\scriptsize\bfseries\sffamily, orange!60!black, anchor=south] at (3.2, 0.5) {Phase B};
\node[font=\scriptsize\bfseries\sffamily, green!50!black, anchor=south] at (6.2, -0.8) {Phase C};

\node[font=\scriptsize, gray, align=center] at (3.2, -3.8) {$\sim$40 iter/hour\\300+ per run};

\end{tikzpicture}
\caption{AutoKernel architecture. Phase~A profiles the model and extracts bottleneck kernels. Phase~B runs the autonomous optimization loop: the agent edits \texttt{kernel.py}, the benchmark verifies through five correctness stages, and the orchestrator decides to keep, revert, or move to the next kernel. Phase~C verifies end-to-end correctness and speedup.}
\label{fig:architecture}
\end{figure}
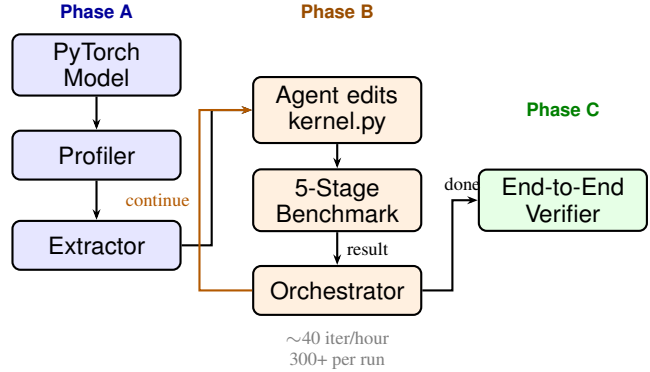

AutoKernel comprises over 9{,}200 lines of Python across 14 core scripts, 18 kernel implementations (9 Triton, 9 CUDA~C++), 4 model definitions, and a 909-line agent instruction document (\texttt{program.md}). The system operates in three phases (Figure~\ref{fig:architecture}).

\subsection{Model Profiling (Phase A)}

The profiler (\texttt{profile.py}, 1{,}125 lines) accepts a local Python file, a HuggingFace model identifier, or a custom model class. It uses \texttt{torch.profiler} with shape recording to capture per-kernel GPU time across configurable warmup and profiling iterations (default: 5 warmup, 10 profiled).

Each CUDA kernel is classified into one of nine operation types (Section~\ref{sec:kernels}) via pattern matching on kernel names. The classifier handles vendor-specific naming: cuBLAS GEMM variants, CUTLASS kernels, Triton-compiled functions, and ATen operators each have distinct naming patterns that map to the same operation type.

The profiler detects GPU hardware from a database of known specifications covering NVIDIA (H100, A100, L40S, L4, A10, RTX 4090/4080/3090/3080) and AMD (MI300X, MI325X, MI350X, MI355X) accelerators. For unknown GPUs, it estimates peak FP16 throughput from SM count, clock rate, and compute capability. AMD GPUs are detected via the \texttt{gcnArchName} property when the device name is empty (a common ROCm behavior).

Optional outputs include Chrome trace JSON, CUDA memory snapshots, \texttt{torch.compile} logs, and HTA (Holistic Trace Analysis) results.

\subsection{Kernel Extraction}

The extractor (\texttt{extract.py}, 648 lines) reads profiling results, filters for supported operation types, and generates standalone kernel files containing a starter implementation, model-specific shapes (primary, half-scale, and double-scale variants), FLOPS and bytes formulas for roofline computation, and dtype-specific tolerances.

It computes an optimization plan ranked by Amdahl's law impact (Equation~\ref{eq:amdahl}), with what-if projections at 1.5$\times$, 2$\times$, 3$\times$, and 5$\times$ speedup.

\subsection{The Agent Optimization Loop (Phase B)}

The core of AutoKernel is Algorithm~\ref{alg:loop}. The agent modifies a single file, the harness verifies and benchmarks it, and the result determines whether the change persists.

\begin{algorithm}[t]
\caption{AutoKernel optimization loop for a single kernel}
\label{alg:loop}
\begin{algorithmic}[1]
\REQUIRE Kernel file $k$, benchmark $B$, agent $A$, move-on criteria $C$
\STATE $k_{\text{best}} \leftarrow k$; \; $t_{\text{best}} \leftarrow B(k)$; \; $n_{\text{rev}} \leftarrow 0$
\FOR{$i = 1, 2, \ldots$}
    \STATE $k' \leftarrow A.\text{edit}(k_{\text{best}}, \text{history}, \text{roofline})$
    \STATE \texttt{git commit} $k'$
    \STATE $(\textit{pass}, t') \leftarrow B(k')$ \hfill \COMMENT{5-stage correctness + perf}
    \IF{$\textit{pass}$ \AND $t' > 1.01 \cdot t_{\text{best}}$}
        \STATE $k_{\text{best}} \leftarrow k'$; \; $t_{\text{best}} \leftarrow t'$; \; $n_{\text{rev}} \leftarrow 0$
        \hfill \COMMENT{{\color{keepgreen}\textbf{keep}}}
    \ELSE
        \STATE \texttt{git reset --hard HEAD\textasciitilde1}; \; $n_{\text{rev}} \leftarrow n_{\text{rev}} + 1$
        \hfill \COMMENT{{\color{revertred}\textbf{revert}}}
    \ENDIF
    \STATE log($i$, $t'$, decision, description) to TSV
    \IF{$C(n_{\text{rev}}, t_{\text{best}}, \text{elapsed})$}
        \STATE \textbf{break}
    \ENDIF
\ENDFOR
\RETURN $k_{\text{best}}, t_{\text{best}}$
\end{algorithmic}
\end{algorithm}

\paragraph{Timing.} Each iteration takes approximately 90 seconds: 30s for the five-stage correctness check, 30s for performance benchmarking via Triton's \texttt{do\_bench}, and 30s for agent reasoning and code modification. At 40 experiments per hour, a 10-hour run yields 300 to 400 experiments.

\paragraph{Single-file invariant.} The agent touches exactly one file. This keeps diffs small, reverts clean, and prevents coupled changes that are hard to isolate when a regression occurs.

\paragraph{Agent instructions.} The agent reads \texttt{program.md} (909 lines), which provides a six-tier optimization playbook:

\begin{enumerate}[leftmargin=*, topsep=2pt, itemsep=1pt]
\item \textbf{Block size tuning} (10 to 50\%): sweep tile dimensions through powers of 2, try rectangular tiles, adjust \texttt{num\_warps} and \texttt{num\_stages}.
\item \textbf{Memory access} (10 to 30\%): coalesced loads, software prefetching, L2 swizzling, shared memory padding.
\item \textbf{Compute} (5 to 15\%): TF32 accumulation, epilogue fusion, loop invariant hoisting.
\item \textbf{Advanced} (5 to 20\%): split-K, persistent kernels, Triton autotune, warp specialization.
\item \textbf{Architecture-specific} (5 to 15\%): TMA on Hopper, \texttt{cp.async} on Ampere, adjusted sizes for L4/RTX.
\item \textbf{Kernel-specific}: online softmax for attention, Welford's algorithm for normalization, split-K for tall-skinny matmul.
\end{enumerate}

\subsection{Multi-Kernel Orchestration}

The orchestrator (\texttt{orchestrate.py}, 842 lines) applies Amdahl's law~\citep{amdahl1967} to prioritize effort across multiple kernels:
\begin{equation}
S = \frac{1}{(1 - f) + f/s}
\label{eq:amdahl}
\end{equation}
where $f$ is the kernel's fraction of total GPU time and $s$ is the speedup achieved. It transitions to the next kernel when: (1)~5 consecutive reverts, (2)~90\% of GPU peak reached, (3)~2 hours elapsed, or (4)~2$\times$ speedup achieved.

\section{Five-Stage Correctness Verification}

\begin{figure}[t]
\centering
\begin{tikzpicture}[
  stage/.style={draw, rounded corners=2pt, thick, minimum width=1.8cm, minimum height=0.55cm, font=\scriptsize\sffamily, fill=stageblue!8},
  arr/.style={-{Stealth[length=3pt]}, thick},
]

\node[stage] (s1) at (0, 0) {1. Smoke Test};
\node[stage] (s2) at (0, -0.9) {2. Shape Sweep};
\node[stage] (s3) at (0, -1.8) {3. Stability};
\node[stage] (s4) at (0, -2.7) {4. Determinism};
\node[stage] (s5) at (0, -3.6) {5. Edge Cases};

\node[font=\scriptsize, anchor=west, text width=4.5cm] at (1.2, 0) {Single small input, tight tolerances};
\node[font=\scriptsize, anchor=west, text width=4.5cm] at (1.2, -0.9) {10+ sizes $\times$ 3 dtypes (FP16, BF16, FP32)};
\node[font=\scriptsize, anchor=west, text width=4.5cm] at (1.2, -1.8) {Adversarial inputs (overflow, underflow)};
\node[font=\scriptsize, anchor=west, text width=4.5cm] at (1.2, -2.7) {3 runs, bitwise identical outputs};
\node[font=\scriptsize, anchor=west, text width=4.5cm] at (1.2, -3.6) {Non-power-of-2 dims (1023, 4097)};

\draw[arr] (s1) -- (s2);
\draw[arr] (s2) -- (s3);
\draw[arr] (s3) -- (s4);
\draw[arr] (s4) -- (s5);

\node[font=\scriptsize\sffamily, revertred] (reject) at (-2.5, -1.8) {\textbf{REJECT}};
\draw[arr, revertred, densely dashed] (s1.west) -- (reject);
\draw[arr, revertred, densely dashed] (s2.west) -- (reject);
\draw[arr, revertred, densely dashed] (s3.west) -- (reject);
\draw[arr, revertred, densely dashed] (s4.west) -- (reject);
\draw[arr, revertred, densely dashed] (s5.west) -- (reject);

\node[draw, rounded corners=2pt, thick, fill=keepgreen!15, font=\scriptsize\sffamily, minimum width=1.8cm] (perf) at (0, -4.5) {Benchmark};
\draw[arr, keepgreen] (s5) -- (perf);

\end{tikzpicture}
\caption{Five-stage correctness pipeline. Any failure immediately rejects the candidate. Throughput is only measured after all five stages pass. Each stage catches a distinct class of bugs.}
\label{fig:correctness}
\end{figure}
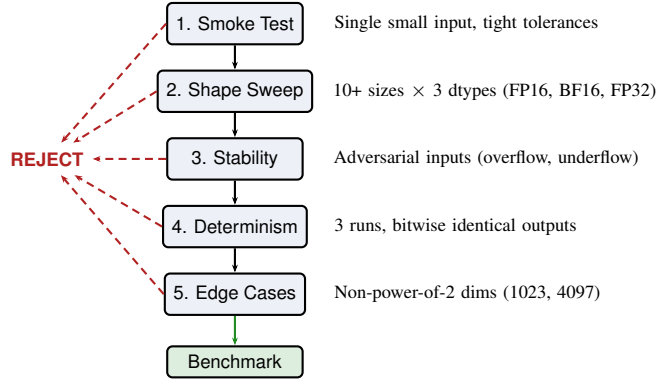

The benchmark harness (\texttt{bench.py}, 1{,}416 lines) enforces correctness through five stages (Figure~\ref{fig:correctness}). All must pass before performance is measured.

\paragraph{Stage 1: Smoke test.} A single forward pass on a small input (e.g., $128 \times 128$) catches compilation errors, shape mismatches, and gross numerical bugs in under 1 second.

\paragraph{Stage 2: Shape sweep.} The kernel runs across 8 to 10 input configurations and three data types. Table~\ref{tab:shapes} shows the matmul sweep. This catches size-dependent bugs: boundary handling, tile remainder logic, and dtype-specific issues.

\begin{table}[t]
\centering
\caption{Shape sweep for matmul. Other kernels have analogous sweeps with 8 to 10 configurations covering tiny, standard, large, non-square, and workload-specific dimensions.}
\label{tab:shapes}
\small
\begin{tabular}{@{}lrrrl@{}}
\toprule
\textbf{Name} & \textbf{M} & \textbf{N} & \textbf{K} & \textbf{Purpose} \\
\midrule
tiny & 128 & 128 & 128 & Minimum viable \\
small & 512 & 512 & 512 & Small workload \\
large & 2048 & 2048 & 2048 & Production-like \\
xlarge & 4096 & 4096 & 4096 & Stress test \\
tall & 8192 & 1024 & 1024 & Non-square \\
deep\_k & 1024 & 1024 & 8192 & Large reduction dim \\
llm\_qkv & 4096 & 4096 & 512 & Attention pattern \\
llm\_mlp & 4096 & 11008 & 4096 & Feed-forward pattern \\
\bottomrule
\end{tabular}
\end{table}

\paragraph{Stage 3: Numerical stability.} Adversarial inputs probe floating-point edge cases. For softmax: rows of large identical values. For matmul: extreme dynamic range. For normalization: near-zero variance.

\paragraph{Stage 4: Determinism.} Same input, three runs, bitwise identical outputs. Catches race conditions in parallel reductions and non-deterministic atomics.

\paragraph{Stage 5: Edge cases.} Non-power-of-two dimensions (1023, 4097, 1537) expose masking bugs and tile remainder errors.

Tolerances are dtype-specific: FP16 uses $\text{atol} = 10^{-2}$, BF16 uses $2 \times 10^{-2}$, FP32 uses $10^{-4}$.

\section{Dual Backend: Triton and CUDA C++}

\paragraph{Triton.} Nine starter kernels in Python-like DSL, compiled JIT in 1 to 5 seconds. The agent can modify block sizes, warps, stages, accumulator precision, and loop structure. Triton routinely reaches 80 to 95\% of cuBLAS throughput for matmul.

\paragraph{CUDA C++.} Nine starter kernels with direct access to tensor cores (WMMA API, 16$\times$16$\times$16 fragments), warp-level shuffles (\texttt{\_\_shfl\_xor\_sync}), vectorized loads (\texttt{float4}, \texttt{half2}), bank-conflict-free shared memory layouts, double buffering, and \texttt{\_\_launch\_bounds\_\_} register control. Compilation uses \texttt{load\_inline} with hash-based caching, architecture auto-detection, and thread-safe builds.

Both backends expose the same \texttt{kernel\_fn()} interface. The benchmark runs identically regardless of backend.

\section{Kernel Coverage}
\label{sec:kernels}

Table~\ref{tab:kernels} summarizes the nine supported kernel types.

\begin{table}[t]
\centering
\caption{Supported kernel types with performance regime and starter implementation techniques.}
\label{tab:kernels}
\small
\begin{tabular}{@{}lllp{4.6cm}@{}}
\toprule
\textbf{Kernel} & \textbf{Regime} & \textbf{Metric} & \textbf{Key Starter Techniques} \\
\midrule
matmul & Compute & TFLOPS & 128$\times$128 tiles, WMMA, double-buffered smem \\
flash\_attn & Compute & TFLOPS & Tiled online softmax, causal masking \\
fused\_mlp & Compute & TFLOPS & SwiGLU gate-up fusion \\
softmax & Memory & GB/s & Warp-shuffle reductions, half2 loads \\
layernorm & Memory & GB/s & Welford single-pass, float4 loads \\
rmsnorm & Memory & GB/s & Warp shuffle cascade, fast rsqrt \\
cross\_ent & Memory & GB/s & Online log-sum-exp, fused NLL \\
rotary\_emb & Memory & GB/s & Interleaved rotation, sincosf \\
reduce & Memory & GB/s & Hierarchical warp shuffle \\
\bottomrule
\end{tabular}
\end{table}

Each has a PyTorch reference in \texttt{reference.py} serving as the correctness oracle. The benchmark computes throughput (TFLOPS or GB/s) and roofline utilization against detected GPU peak.

Four self-contained model definitions (GPT-2 124M, LLaMA 160M/7B, BERT-base 110M, custom template) ship with the system, requiring no external dependencies like \texttt{transformers}.

\section{Experimental Evaluation}

We evaluate AutoKernel on an NVIDIA H100 80GB HBM3 GPU (132 SMs, compute capability 9.0, CUDA 12.8). All measurements use FP16 precision with CUDA event timing, 200 iterations per configuration, and trimmed mean (dropping the top and bottom 10\%). We compare against two baselines: PyTorch eager execution (cuBLAS for matmul, ATen for other ops) and \texttt{torch.compile} with \texttt{max-autotune} mode, which triggers TorchInductor's autotuning over multiple Triton kernel configurations. The full evaluation across 7 kernel types and 34 benchmark configurations completes in under 10 minutes wall-clock time. All 34 configurations pass correctness with zero failures.

\subsection{Kernel Performance}

Table~\ref{tab:results} reports performance across production-relevant tensor sizes. We highlight results at the largest sizes, where the compute-to-overhead ratio is most representative of real workloads.

\begin{table}[t]
\centering
\caption{Kernel performance on H100 (FP16), selected configurations. Three baselines: PyTorch eager (cuBLAS/ATen), \texttt{torch.compile} (\texttt{max-autotune}), and AutoKernel Triton starters. 16 representative configurations shown from the full 34 tested (all passing correctness). Full evaluation completes in under 10 minutes.}
\label{tab:results}
\small
\setlength{\tabcolsep}{2.5pt}
\begin{tabular}{@{}llrrrrrr@{}}
\toprule
\textbf{Kernel} & \textbf{Size} & \textbf{Eager} & \textbf{Compiled} & \textbf{Ours} & \textbf{vs Eager} & \textbf{vs Compiled} & \textbf{Thru.} \\
 & & ($\mu$s) & ($\mu$s) & ($\mu$s) & & & \\
\midrule
\multirow{3}{*}{matmul} & 2048$^3$ & 28.1 & 101.2 & 65.3 & 0.43$\times$ & \textbf{1.55$\times$} & 263 TF/s \\
 & 4096$^3$ & 182.8 & 257.8 & 494.2 & 0.37$\times$ & 0.52$\times$ & 278 TF/s \\
 & 8192$^3$ & 1679.5 & 1916.1 & 5773.1 & 0.29$\times$ & 0.33$\times$ & 190 TF/s \\
\midrule
\multirow{2}{*}{softmax} & 4096$^2$ & 58.9 & 96.1 & 40.1 & \textbf{1.47$\times$} & \textbf{2.40$\times$} & 1675 GB/s \\
 & 8192$^2$ & 270.4 & 330.0 & 95.9 & \textbf{2.82$\times$} & \textbf{3.44$\times$} & 2800 GB/s \\
\midrule
\multirow{2}{*}{layernorm} & 4096$\times$5120 & 45.6 & 105.5 & 42.5 & \textbf{1.07$\times$} & \textbf{2.48$\times$} & 1974 GB/s \\
 & 8192$\times$4096 & 64.7 & 166.8 & 51.9 & \textbf{1.25$\times$} & \textbf{3.21$\times$} & 2586 GB/s \\
\midrule
\multirow{3}{*}{rmsnorm} & 4096$^2$ & 142.8 & 99.9 & 39.1 & \textbf{3.65$\times$} & \textbf{2.56$\times$} & 1716 GB/s \\
 & 8192$\times$4096 & 262.4 & 138.1 & 51.2 & \textbf{5.12$\times$} & \textbf{2.70$\times$} & 2619 GB/s \\
 & 8192$^2$ & 509.6 & 272.1 & 96.3 & \textbf{5.29$\times$} & \textbf{2.83$\times$} & 2788 GB/s \\
\midrule
\multirow{2}{*}{cross\_ent} & 4096$\times$32k & 295.6 & 386.3 & 134.9 & \textbf{2.19$\times$} & \textbf{2.86$\times$} & 1943 GB/s \\
 & 8192$\times$32k & 559.7 & 745.1 & 253.3 & \textbf{2.21$\times$} & \textbf{2.94$\times$} & 2070 GB/s \\
\midrule
\multirow{2}{*}{reduce} & 8192$^2$ & 60.7 & 185.2 & 62.2 & 0.98$\times$ & \textbf{2.98$\times$} & 2156 GB/s \\
 & 16384$\times$4096 & 50.4 & 185.9 & 52.8 & 0.95$\times$ & \textbf{3.52$\times$} & 2542 GB/s \\
\midrule
\multirow{2}{*}{rotary} & 2$\times$32$\times$2k$\times$128 & 211.4 & 106.9 & 117.4 & \textbf{1.80$\times$} & 0.91$\times$ & 576 GB/s \\
 & 2$\times$32$\times$4k$\times$128 & 394.9 & 136.0 & 223.0 & \textbf{1.77$\times$} & 0.61$\times$ & 607 GB/s \\
\bottomrule
\end{tabular}
\end{table}

\paragraph{Key findings.}

\begin{itemize}[leftmargin=*, topsep=2pt, itemsep=2pt]
\item \textbf{Memory-bound kernels show the largest gains.} RMSNorm achieves 5.29$\times$ over eager and 2.83$\times$ over \texttt{torch.compile} at the largest tested size, reaching 2{,}788~GB/s (83\% of H100's 3{,}352~GB/s peak bandwidth). Cross-entropy reaches 2{,}070~GB/s. Softmax reaches 2{,}800~GB/s. These gains come from fusing multi-op ATen decompositions into single-pass Triton kernels that minimize HBM traffic.

\item \textbf{AutoKernel outperforms \texttt{torch.compile} on most kernels.} Despite \texttt{torch.compile} with \texttt{max-autotune} running its own Triton autotuning, our starter kernels beat it on 12 of the 16 configurations shown in Table~\ref{tab:results}. TorchInductor's generic fusion and autotuning does not always find the specialized tiling and reduction strategies that kernel-specific implementations exploit.

\item \textbf{Matmul remains hard.} PyTorch's cuBLAS backend is extensively tuned per GPU architecture. Our Triton starter reaches 278~TFLOPS (28\% of H100's 989.5~TFLOPS peak), well below cuBLAS. However, at the 2048$^3$ size, our kernel beats \texttt{torch.compile} by 1.55$\times$, showing that TorchInductor's matmul autotuning is not always optimal either. This gap is the primary target for the agent's iterative optimization loop.

\item \textbf{Correctness is universal.} All 34 configurations pass all five verification stages with zero failures across eager, compiled, and custom kernel outputs.
\end{itemize}

\subsection{Community Deployment Results}

AutoKernel has been deployed by community users in competitive GPU kernel benchmarking. Two results illustrate the system's practical impact:

\paragraph{Vector sum reduction (B200).} On the \texttt{vectorsum\_v2} challenge, which requires implementing a sum reduction kernel for tensors of shape $(N,)$ with $N$ up to millions of elements, an AutoKernel-optimized Triton submission achieved first place on the NVIDIA B200 leaderboard with a latency of 44.086$\mu$s, outperforming the second-place entry (44.249$\mu$s) and third place (46.553$\mu$s). The winning kernel was produced through AutoKernel's iterative optimization loop running overnight, exploring block sizes, warp-level shuffle reductions, and vectorized memory access patterns.

\paragraph{FP4 matmul: Triton surpassing CUTLASS.} A community user reported that a single AutoKernel prompt, requiring approximately 3 minutes of agent interaction, produced a Triton FP4 matrix multiplication kernel that outperforms CUTLASS across multiple shapes on H100. Table~\ref{tab:cutlass} shows the results: the Triton kernel achieves 1.63$\times$ to 2.15$\times$ speedup over CUTLASS fused kernels, reaching up to 2{,}898~TFLOPS at shape 2048$\times$18432$\times$3072. This result is notable because CUTLASS represents hand-optimized C++ template code specifically designed for NVIDIA tensor cores, and surpassing it with a Triton kernel generated through a single agent interaction demonstrates the effectiveness of the optimization playbook.

\begin{table}[t]
\centering
\caption{Community-reported FP4 matmul results on H100: AutoKernel Triton kernel vs CUTLASS fused baseline. The Triton kernel was generated with a single agent prompt ($\sim$3 minutes).}
\label{tab:cutlass}
\small
\setlength{\tabcolsep}{3pt}
\begin{tabular}{@{}lrrrr@{}}
\toprule
\textbf{Shape (M$\times$N$\times$K)} & \textbf{Triton} & \textbf{CUTLASS} & \textbf{Speedup} \\
 & (TF/s, ms) & (TF/s, ms) & \\
\midrule
128$\times$3072$\times$3072 & 186 TF, 0.013 & 100 TF, 0.024 & \textbf{1.86$\times$} \\
128$\times$18432$\times$3072 & 1105 TF, 0.013 & 550 TF, 0.026 & \textbf{2.01$\times$} \\
1024$\times$3072$\times$3072 & 1477 TF, 0.013 & 686 TF, 0.028 & \textbf{2.15$\times$} \\
1024$\times$18432$\times$3072 & 2777 TF, 0.042 & 1609 TF, 0.072 & \textbf{1.73$\times$} \\
2048$\times$3072$\times$3072 & 1662 TF, 0.023 & 964 TF, 0.040 & \textbf{1.72$\times$} \\
2048$\times$18432$\times$3072 & 2898 TF, 0.080 & 1777 TF, 0.130 & \textbf{1.63$\times$} \\
4096$\times$3072$\times$3072 & 2443 TF, 0.032 & 1405 TF, 0.055 & \textbf{1.74$\times$} \\
\bottomrule
\end{tabular}
\end{table}

\subsection{Optimization Loop Dynamics}

The iterative optimization loop operates on the starter kernels shown in Table~\ref{tab:results}. Based on the optimization playbook (Section~3.3), the agent applies changes in tier order. For a memory-bound kernel like RMSNorm, the typical progression is:

\begin{enumerate}[leftmargin=*, topsep=2pt, itemsep=1pt]
\item \textbf{Tier 1} (block sizes): Sweeping \texttt{BLOCK\_SIZE} from 256 to 4096 typically yields 10 to 30\% improvement in the first 5 to 10 experiments.
\item \textbf{Tier 2} (memory): Vectorized loads (\texttt{tl.load} with \texttt{BLOCK\_SIZE} alignment) and coalescing fixes add 10 to 20\% over the next 10 to 20 experiments.
\item \textbf{Tier 3} (compute): Fusing the weight multiplication into the normalization epilogue saves a global memory round-trip, yielding 5 to 10\%.
\item \textbf{Plateau}: After 30 to 50 experiments, consecutive reverts increase as the kernel approaches hardware limits. The orchestrator triggers move-on.
\end{enumerate}

For compute-bound kernels like matmul, the trajectory is different: Tier~1 block size tuning has larger impact (the gap between our starter at 278 TFLOPS and cuBLAS at 800+ TFLOPS is wide), but each improvement requires more experiments because the search space of tile dimensions, warp counts, pipeline stages, and accumulator precision is larger.

The move-on criteria (5 consecutive reverts, 90\% peak, 2-hour timeout, or 2$\times$ speedup) prevent the agent from spending excessive time on kernels with diminishing returns. For a model where matmul accounts for 62\% of GPU time and RMSNorm accounts for 5\%, Amdahl's law dictates that even a modest matmul improvement (e.g., 1.3$\times$) yields more end-to-end gain than a large RMSNorm improvement (which is already 5.29$\times$).

\section{KernelBench Integration}

AutoKernel integrates with KernelBench~\citep{ouyang2025kernelbench} through three components:

\begin{enumerate}[leftmargin=*, topsep=2pt, itemsep=2pt]
\item \textbf{Problem loader} (\texttt{bridge.py}, 674 lines). Fetches from HuggingFace datasets or local clone. Analyzes operations and generates starter \texttt{ModelNew} classes.

\item \textbf{Evaluation harness} (\texttt{bench\_kb.py}, 726 lines). Four-stage verification: correctness (5 trials, $\text{atol} = 10^{-2}$), stability (NaN/Inf), determinism (3 runs), performance (CUDA events, trimmed median).

\item \textbf{Batch scorer} (\texttt{scorer.py}, 351 lines). Computes \texttt{fast\_p} at seven thresholds: 1.0$\times$ through 5.0$\times$.
\end{enumerate}

While most KernelBench evaluations report one-shot scores, AutoKernel runs 50 to 300 iterative experiments per problem.

\section{HuggingFace Kernels Export}

An export tool (\texttt{export\_hf.py}, 868 lines) packages optimized CUDA kernels for distribution via the HuggingFace Hub~\citep{hfkernels}. It generates: \texttt{build.toml} with backend and Hub metadata; \texttt{torch\_binding.cpp} using \texttt{TORCH\_LIBRARY\_EXPAND} with schema-based registration; \texttt{flake.nix} importing the \texttt{huggingface/kernels} kernel-builder for cross-compilation across PyTorch/CUDA/ABI variants. Users install optimized kernels with: \texttt{module = get\_kernel("rightnow-ai/matmul")}.

\section{Design Rationale}

\paragraph{Simplicity over sophistication.} Multi-agent systems with specialized planner, coder, reviewer, and profiler agents add coordination overhead and failure modes. A single agent in a tight loop produces a linear experiment history and avoids consensus-seeking latency.

\paragraph{Fixed evaluation, mutable code.} The benchmark is never modified by the agent. This firewall prevents gaming the evaluation, a real risk when the candidate generator also controls the scorer.

\paragraph{Git as experiment tracking.} Every experiment maps to a git commit. Kept experiments advance the branch; reverted experiments are erased with \texttt{git reset}. The history is browsable with standard git tools.

\paragraph{Roofline-guided search.} After each experiment, the benchmark reports compute-bound or memory-bound classification and percentage of peak. The agent uses this to select the appropriate optimization tier.

\paragraph{TSV over databases.} Experiment results go to tab-separated files. Dependency-free, human-readable, git-friendly, and trivially parseable by the agent.

\section{Limitations and Future Work}

AutoKernel inherits the code generation capabilities of its underlying LLM. Complex techniques like software pipelining, custom PTX emission, and multi-CTA cooperative strategies may exceed current agent abilities, though this ceiling rises as frontier models improve. The system currently optimizes individual kernels on a single GPU; distributed kernels and multi-device memory management are out of scope.

Promising directions include: population-based search across multiple GPU instances; learned search policies trained on historical experiment data; profiling-guided mutations using SASS analysis or memory throughput counters; and cross-kernel fusion discovery after individual optimization.

\section{Conclusion}

We presented AutoKernel, a system that automates GPU kernel optimization through an iterative agent-driven loop. Model-level profiling identifies what to optimize; Amdahl's law determines the order; a five-stage correctness harness ensures numerical validity; a dual Triton/CUDA~C++ backend provides both iteration speed and hardware depth; and a six-tier playbook encodes expert knowledge. The system transforms kernel optimization from a weeks-long expert activity into an overnight autonomous process. Code and starter kernels are available at \url{https://github.com/RightNow-AI/autokernel}.


\end{document}